\documentclass[runningheads]{llncs}
\usepackage{graphicx}
\usepackage{amsmath,amssymb} 
\usepackage{color}
\usepackage{verbatim}
\usepackage[width=122mm,left=12mm,paperwidth=146mm,height=193mm,top=12mm,paperheight=217mm]{geometry}

\usepackage{multirow}
\usepackage{changepage}
\usepackage{caption}
\usepackage{subcaption}
\begin{document}
\pagestyle{headings}
\mainmatter

\title{Learning Models for Actions and Person-Object Interactions with Transfer to Question Answering} 

\titlerunning{Learning Models for Actions and Person-Object Interactions with Transfer to Question Answering}

\authorrunning{Arun Mallya and Svetlana Lazebnik}

\author{Arun Mallya and Svetlana Lazebnik}


\institute{
  University of Illinois at Urbana-Champaign\\
  \email{ \{amallya2,slazebni\}@illinois.edu}
}

\maketitle

\begin{abstract}

This paper proposes deep convolutional network models that utilize local and global context to make human activity label predictions in still images, achieving state-of-the-art performance on two recent datasets with hundreds of labels each. We use multiple instance learning to handle the lack of supervision on the level of individual person instances, and weighted loss to handle unbalanced training data. Further, we show how specialized features trained on these datasets can be used to improve accuracy on the Visual Question Answering (VQA) task, in the form of multiple choice fill-in-the-blank questions (Visual Madlibs). Specifically, we tackle two types of questions on person activity and person-object relationship and show improvements over generic features trained on the ImageNet classification task.





\end{abstract}

\section{Introduction}
\label{sec:intro}

The task of Visual Question Answering (VQA) has recently garnered a lot of interest with multiple datasets~\cite{malinowski2014multi,VQA,yu2015visual} and systems~\cite{xu2015ask,zhou2015simple,gao2015you,shih2015look,andreas2016neural,fukui2016multimodal,jabri2016revisiting} being proposed. Many of these systems rely on features extracted from deep convolutional neural networks (CNNs) pre-trained on the ImageNet classification task~\cite{ILSVRC15}, with or without fine-tuning on the VQA dataset at hand. However, questions in VQA datasets tend to cover a wide variety of concepts such as the presence or absence of objects, counting, brand name recognition, emotion, activity, scene recognition and more. Generic ImageNet-trained networks are insufficiently well tailored for such open-ended tasks, and the VQA datasets themselves are currently too small to provide adequate training data for all types of visual content that are covered in their questions.

\begin{figure}[h!]
    \centering
    \includegraphics[width=\linewidth]{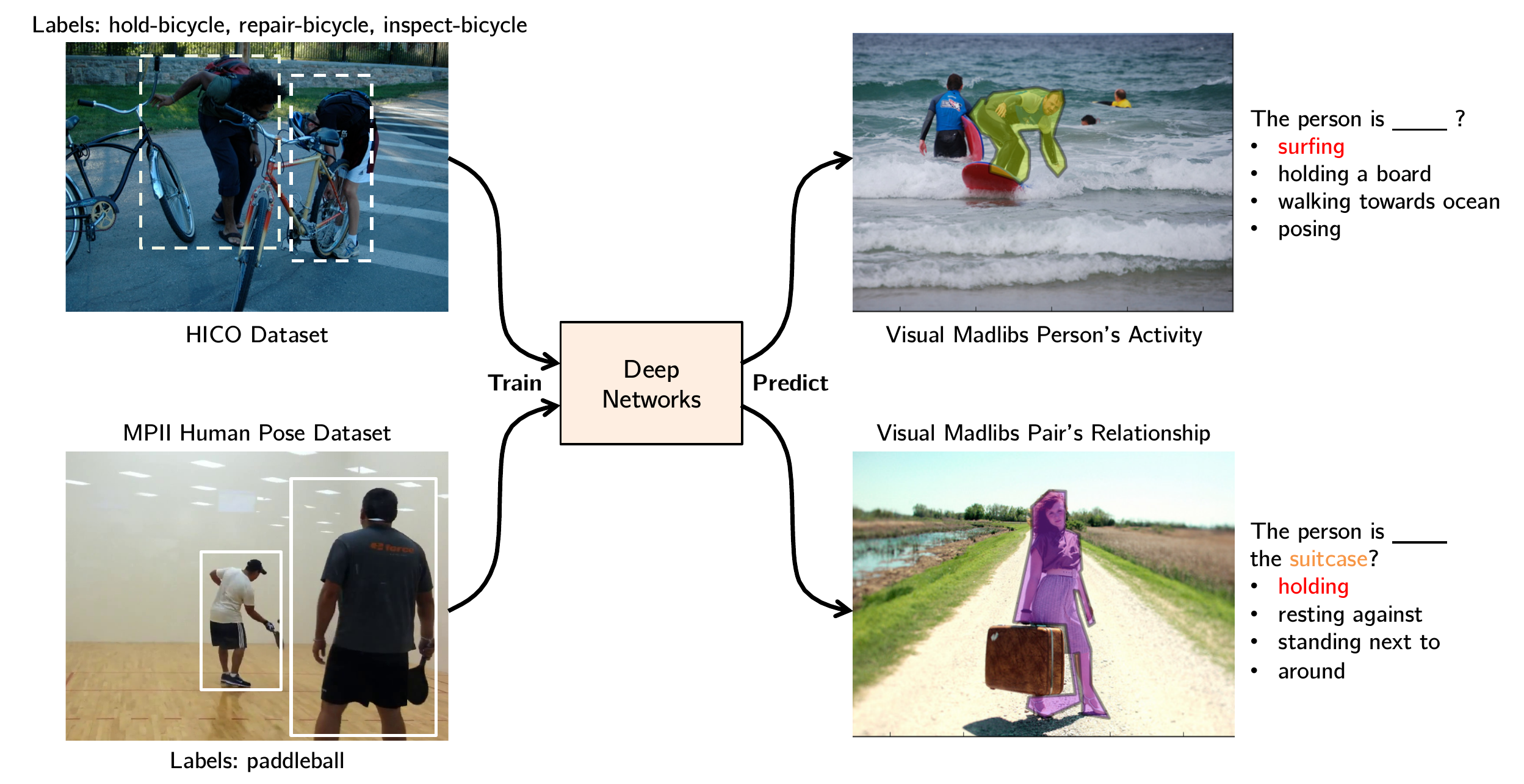}
    \caption{We train CNNs on the HICO and MPII datasets to predict human activity labels. Our networks fuse features from the full image and the person bounding boxes, which are provided in the MPII dataset and detected in the HICO dataset. 
    We then use these networks to answer two types of multiple choice questions from the MadLibs dataset -- about a person's activity, and the relationship between a person and an object.}
    \label{fig:overview}
\end{figure}

Fortunately, we are also seeing the release of valuable datasets targeting specific tasks such as scene recognition~\cite{zhou2014learning}, age, gender, and emotion classification~\cite{levi2015age,levi2015emotion}, human action recognition~\cite{maji2011action,chao2015hico,pishchulin2014fine}, etc. To better understand and answer questions about an image, we should draw on the knowledge from these specialized datasets. Given a specific question type, we should be able to choose features from appropriate expert models or networks. 
In this paper, we show that transferring expert knowledge from a network trained on human activity prediction can not only improve question answering performance, but also help interpret the model's decisions. We train deep networks on the HICO~\cite{chao2015hico} and MPII~\cite{pishchulin2014fine} datasets to predict human activity labels and apply these networks to answer two types of multiple choice fill-in-the-blank questions from the Madlibs dataset~\cite{yu2015visual} on person activity and person-object relationships (Figure~\ref{fig:overview}). Our contributions are as follows:
\begin{enumerate}
\item We propose simple CNN models for predicting human activity labels by fusing features from a person bounding box and global context from the whole image. At training time, the person boxes are provided in the MPII dataset and must be automatically detected in HICO. Our CNN architecture is described in Section~\ref{sec:method}. 
\item At training time, we use Multiple Instance Learning (MIL) to handle the lack of full person instance-label supervision and weighted loss to handle the unbalanced training data. The resulting models beat the previous state-of-the-art on the respective datasets, as shown in Section~\ref{subsec:activity_pred}.
\item We transfer our models to VQA with the help of a standard image-text embedding (canonical correlation analysis or CCA) and show improved accuracy on MadLibs activity and person-object interaction questions in Section~\ref{subsec:mcq}.
\end{enumerate}

\vspace{-5mm}


\section{Related Work}
\label{related_work}

There exist many datasets for action recognition in still images, including the older PASCAL VOC~\cite{everingham2010pascal} and Stanford 40 Actions~\cite{yao2011human}, and newer MPII Human Pose Dataset~\cite{pishchulin2014fine}, COCO-A~\cite{ronchi2015COCOA} and \emph{Humans Interacting with Common Objects} (HICO) dataset~\cite{chao2015hico}. The number of actions in some of the newer datasets is an order of magnitude larger than in the older ones, allowing us to learn vocabularies fit for general VQA. The HICO dataset is currently the largest, consisting of nearly 50000 images belonging to 600 human-object interaction categories. Each category in the HICO dataset is composed of a verb-object pair, with objects belonging to the 80 object categories from the MS COCO dataset~\cite{lin2014microsoft}. On the other hand, the MPII dataset comprises humans performing 393 different activities including walking, running, skating, etc.\ in which they do not necessarily interact with objects. In this work, we train CNNs with simple architectures on HICO and MPII datasets, and show that they outperform the previous state-of-the-art models.

One limitation of the HICO dataset is that it provides labels for the image as a whole, instead of associating them with specific ground truth person instances. We disambiguate activity label assignment over the people in the image with the help of \emph{Multiple Instance Learning} (MIL)~\cite{maron1998MIL}, which has been widely used for recognition problems with weakly or incompletely labeled training data~\cite{zhang2005multiple,hoffman2015detector,vezhnevets2010towards,pinheiro2015image}. In the MIL framework, instead of receiving a set of individually labeled `instances', the learner receives a set of `bags,' each of which is labeled negative if all the instances inside it are negative, and labeled positive if it contains at least one positive instance. In this work, we treat each person bounding box as an `instance' and the image, which contains one or more people in it, as a `bag'. The exact formulation of our learning procedure is explained in Section~\ref{subsec:MIL}.

To recognize a person's activity, we want to use not only the evidence from that person's bounding box, but also some kind of broader contextual information from the image. Previous work suggests the use of latent context boxes~\cite{gkioxari2015rstarcnn}, multiresolution or zoom-out features~\cite{bell2015InsideOutside,mostajabi2015feedforward} and complex 2-D recurrent structures~\cite{bell2015InsideOutside}. In particular, Gkioxari \emph{et al.}~\cite{gkioxari2015rstarcnn} have recently proposed an R$^*$CNN network that chooses a second latent box that overlaps the bounding box of the person and provides the strongest evidence of a particular action being performed. They also proposed a simpler model, the Scene-RCNN, that uses the entire image instead of a chosen box.
We explored using latent boxes but found their performance to be lacking on datasets with hundreds of labels, possibly due to overfitting and the infeasibility of thoroughly sampling latent boxes during training. Similarly, we could not obtain good results with multiresolution features owing to overfitting. Instead, we get surprisingly good results with a simpler architecture combining features from the entire image and the bounding box of the person under consideration, outperforming both R$^*$CNN and Scene-RCNN. 



\section{Action Recognition Method}
\label{sec:method}

\subsection{Network Architecture}
\label{subsec:context}

\begin{figure}[h!]
    \centering
    
    \begin{subfigure}[t]{0.49\linewidth}
        \centering
        \includegraphics[width=0.9\linewidth]{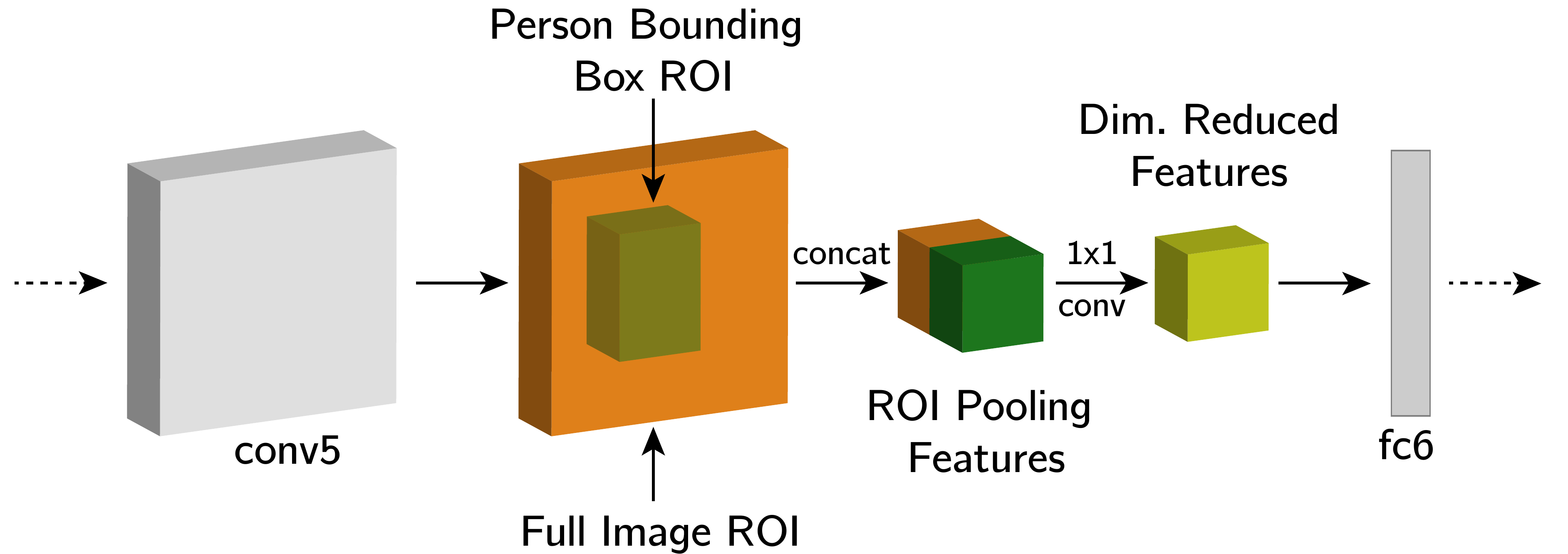}
        \caption{Fusion-1\\}
        \label{fig:network1}
    \end{subfigure}%
    ~ 
    \begin{subfigure}[t]{0.49\linewidth}
        \centering
        \includegraphics[width=\linewidth]{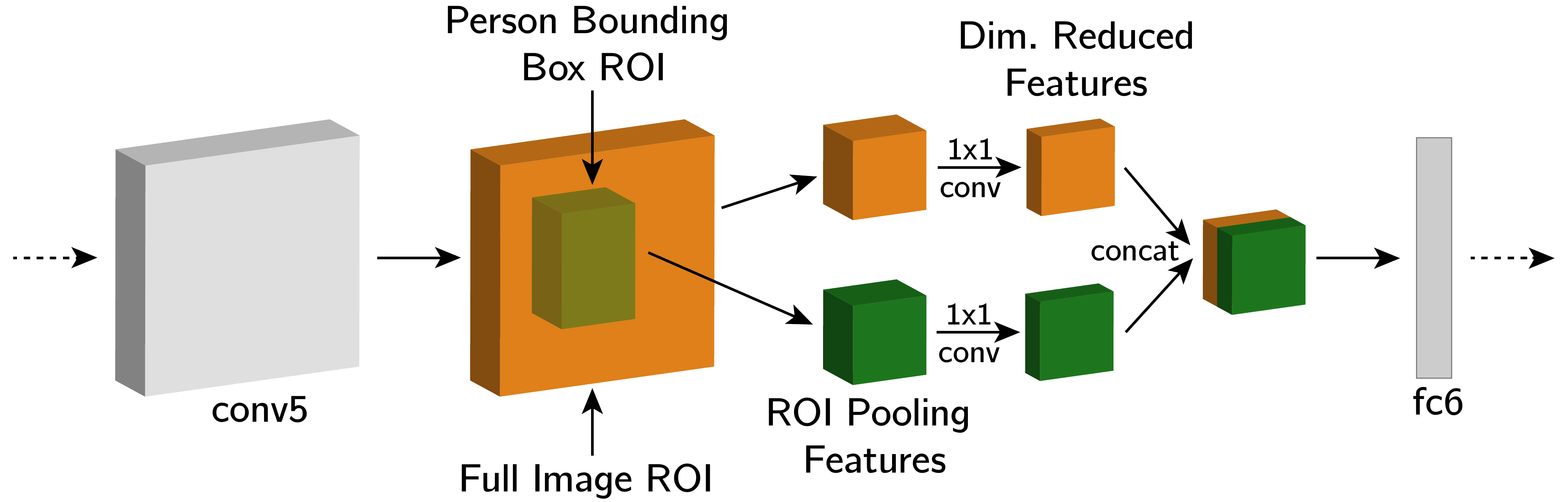}
        \caption{Fusion-2\\}
        \label{fig:network2}
    \end{subfigure}
    \caption{Our networks extract ROI features~\cite{girshick2015fast} of dimension $512\times 7\times 7$ from both the person bounding box and the full image.  The resulting feature is fed into the \emph{fc6} layer of the VGG-16 network. (a) Fusion-1: The two ROI features are stacked and a $1\times 1$ convolution is used for dimensionality reduction. (b) Fusion-2: Each ROI feature is separately reduced using $1\times 1$ convolutions, and the outputs are then stacked.}
    \label{fig:network}
\end{figure}


Our network is based on the \emph{Fast RCNN}~\cite{girshick2015fast} architecture with VGG-16~\cite{simonyan14VGG}. Fast RCNN includes a new adaptive max pooling layer, referred to as the ROI pooling layer, that replaces the standard max pooling layer (\emph{pool5}) after the set of the first five convolutional layers. This layer takes in a list of bounding boxes, referred to as Regions Of Interest (ROI) and outputs a set of fixed-size feature maps for each input ROI that are then fed to the fully connected layers. 
During the forward pass of our network, we use two ROIs for each person instance in the image: the tight bounding box of the person, and the full image (we also experimented with using an expanded person bounding box instead of the full image, but found the full image to always work better). The ROI Pooling layer produces a feature of $512$ channels and spatial size $7\times 7$ for each ROI. 
 The \emph{fc6} layer of the VGG-16 network expects a feature of size $512\times 7\times 7$. 
 
We explore two ways of combining the two ROI features: through stacking and dimensionality reduction (Figure~\ref{fig:network}).
In the first, referred to as Fusion-1, we stack features from the bounding box and the entire image along the channel dimension and obtain a feature of size $1024\times 7\times 7$. A convolutional layer of filter size $1\times 1$ is used to perform dimensionality reduction of channels from $1024$ to $512$, while keeping the spatial size the same.
In the second, referred to as Fusion-2, we first perform dimensionality reduction on the two ROI features individually to reduce the number of channels from $512$ to $256$ each, and then stack the outputs to obtain an input of size $512\times 7\times 7$ for the \emph{fc6} layer.  

Our architecture differs from R$^*$CNN and Scene-RCNN~\cite{gkioxari2015rstarcnn} in two major ways. First, unlike R$^*$CNN, we do not explicitly try to find a box or set of boxes that provide support for a particular label. Second, while R$^*$CNN and Scene-RCNN independently perform prediction using the two features and then average them, we combine features before prediction. The results presented in Section~\ref{subsec:activity_pred} confirm that our ``early'' fusion strategy gives better performance. 
Further, our architecture is faster than R$^*$CNN because it does not need to sample boxes during training and testing.

\subsection{Multiple Instance Learning for Label Prediction}
\label{subsec:MIL}
In the HICO dataset, if at least one of the people in the image is performing an action, the label is marked as positive for the image. As our architecture makes predictions with respect to a person bounding box, we treat the assignment of labels to different people as latent variables and try to infer the assignment during end-to-end training of the network. For an image $I$, let $B$ be the set of all person bounding boxes in the image. Using our network described above which takes as input an image $I$ and a person bounding box $b\in B$, we obtain the score of an action $a$ for the image as follows:
\begin{equation}
    \text{score}(a; I) = \max_{b\in B} \ \text{score}(a; b, I)
\end{equation}
where $\text{score}(a; b, I)$ is the score of action $a$ for the person $b$ in image $I$. The predicted label for the action can be obtained by passing the score through a logistic sigmoid or softmax unit as required. The $\max$ operator enforces the constraint that if a particular action label is active for a given image, then at least one person in the image is performing that action, and when a particular action label is inactive for a given image, then no person in the image is performing the action. During the forward pass, the score and thus the label for the image are predicted using the above relationship. The predicted label is compared to the groundtruth label in order to compute the loss and gradients for backpropagation.


\subsection{Weighted Loss Function}
\label{subsec:wtd_loss}
Mostajabi \emph{et al.}~\cite{mostajabi2015feedforward} showed that use of an asymmetric weighted loss helps greatly in the case of an unbalanced dataset. For the HICO dataset, we have to learn 600 independent classifiers per image and this makes for a highly unbalanced scenario, with the number of negative examples greatly outnumbering the positive examples, even for the most populous categories (an average negative to positive ratio of 6000:1, worst case of 38116:1). We thus compute a weighted cross-entropy loss in which positive examples are weighted by a factor of $w_p$ and negative examples by a factor of $w_n$. Given a training sample $(I, B, y)$ consisting of an image $I$, set of person bounding boxes or detections $B$, and ground truth action label vector $y\in\{0,1\}^C$ for $C$ independent classes, the network produces probabilities of actions being present in the image by passing predictions through a sigmoid activation unit. For any given training sample, the training loss on network prediction $\hat{y}$ is thus given by
\begin{equation}
	\text{loss}(I, B, y) = \sum_{i=1}^{C}  w_p^i \cdot y^i \cdot \log(\hat{y}^i) + w_n^i \cdot (1-y^i) \cdot \log(1-\hat{y}^i)
\end{equation}
In our experiments, we set $w_p=10$ and $w_n=1$ for all classes for simplicity.



\section{Activity Prediction Experiments}
\label{subsec:activity_pred}

{\bf Datasets.} We train and test our system on two different activity classification datasets: HICO~\cite{chao2015hico} and the MPII Human Pose Dataset~\cite{pishchulin2014fine}. The HICO dataset contains labels for 600 human-object interaction activities, any number of which might be simultaneously active for a given image. Labels are provided at the image level even though each image might contain multiple person instances, each performing the same or different activities. The labels can thus be thought of as an aggregate over labels of each person instance in the image. 
As the person bounding boxes are not provided with the HICO dataset, we run the Faster-RCNN detector~\cite{ren2015faster} with the default confidence threshold of 0.8 on all the train and test images. The obtained person bounding boxes are thus not perfect and might have wrong or missing annotations. The HICO training set contains 38,116 images and the test set contains 9,658 images. The training set is highly unbalanced with 51 out of 600 categories having just 1 positive example.

The MPII dataset contains labels for 393 actions. Unlike in HICO, each image only has a single label together with one or more annotated person instances. All person instances inside an image are assumed to be performing the same task. Ground truth bounding boxes are available for each instance in the training set, so we do not need to use MIL can take advantage of the extra training data available by training on each person instance separately. On the test set, however, only a single point inside the bounding box is provided for each instance, so we run the Faster-RCNN detector to detect people. The training set consists of 15,200 images and 22,900 person instances and the test set has 5,709 images. Similar to HICO, the training set is unbalanced and the number of positive examples for a label ranges from 3 to 476 instances.


\begin{table}[h!]
	\begin{center}
		\begin{tabular}{ll|c|c|c|c|c}
			\hline
			& \multicolumn{1}{c|}{\bf Method} & {\bf Full Im.} & {\bf Bbox} & {\bf MIL} & {\bf Wtd. Loss} & {\bf mAP }	\\ \hline
			a) & AlexNet+SVM~\cite{chao2015hico}	& \checkmark & & & & 19.4						\\ \hline
			\multirow{4}{*}{b)} & VGG-16, full image	& \checkmark & & & & 29.4 \\
			& VGG-16, bounding box & & \checkmark & \checkmark &  & 14.6 \\ 
			& VGG-16, R$^\ast$CNN & & \checkmark & \checkmark & & 28.5 		\\ 
			& VGG-16, Scene-RCNN & \checkmark & \checkmark & \checkmark & & 29.0 \\ \hline
			
			\multirow{4}{*}{c)} & Fusion-1 & \checkmark & \checkmark & \checkmark & & 33.6 \\
			& Fusion-1, weighted loss & \checkmark & \checkmark & \checkmark & \checkmark & 36.0 \\

			& Fusion-2 & \checkmark & \checkmark & \checkmark & & 33.8 \\
			& Fusion-2, weighted loss & \checkmark & \checkmark & \checkmark & \checkmark & {\bf 36.1} \\ \hline
			
		\end{tabular}
	\end{center}
	\caption{Performance of various networks on the HICO person-activity dataset. Note that usage of the Bounding Box (Bbox) necessitates the usage of Multiple Instance Learning (MIL).}
	\label{table:hico_results}
\end{table}

{\bf HICO Results.} On the HICO dataset, we compare the networks described in the previous section with VGG-16 networks trained on just the person bounding boxes and just the full image, as well as with R$^\ast$CNN and Scene-RCNN. For the latter two, we use the authors' implementation~\cite{gkioxari2015rstarcnn}. For all the networks, except the R$^\ast$CNN, we use a learning rate of $10^{-5}$, decayed by a factor of 0.1 every 30000 iterations. For the R$^\ast$CNN, we use the recommended setting from~\cite{gkioxari2015rstarcnn} of a learning rate of $10^{-4}$, with a lower and upper intersection over union (IoU) bound for secondary regions of 0.2 and 0.75 and sample 10 secondary regions per person bounding box during a single training pass. We train all networks for 60000 iterations with a momentum of 0.9. Further, all networks are finetuned till the \emph{conv3} layer as in previous work~\cite{gkioxari2015rstarcnn,girshick2015fast}. We use a batch size of 10 images, resize images to a maximum size of 640 pixels, and sample a maximum of 6 person bounding boxes per image in order to fit the network in the GPU memory during training with MIL. Consistent with~\cite{girshick2014rich,agrawal2014analyzing,bell2015InsideOutside}, we initialize our models with weights from the ImageNet-trained VGG-16. 

Table~\ref{table:hico_results} presents our comparison. As HICO is fairly new, the only published baseline~\cite{chao2015hico} uses the AlexNet~\cite{alexnet} (Table~\ref{table:hico_results}a). Using the VGG-16 network improves upon AlexNet by 10 mAP (first line of Table~\ref{table:hico_results}b). The VGG-16 network that uses just the person bounding box to make predictions with MIL performs poorly with only 14.6 mAP (second line of Table~\ref{table:hico_results}b). This is not entirely surprising since the object that the person is interacting with is often not inside that person's bounding box.
More surprisingly, the R$^\ast$CNN architecture, which tries to find secondary boxes to support the person box, performs slightly worse than the full-image VGG network. One possible reason for this is that R$^\ast$CNN has to use MIL twice during training: once for finding the secondary box for an instance, and then again while aggregating over the multiple person instances in the image. Since R$^*$CNN samples only 10 boxes per person instance during each pass of training (same as in~\cite{gkioxari2015rstarcnn}), finding the right box for each of the 600 actions might be difficult. The Scene-RCNN, which uses the entire image as the secondary box, needs to do MIL just once, and performs marginally better than R$^*$CNN. Another possible reason why both R$^*$CNN and Scene-RCNN cannot outperform a full-image network is that they attempt to predict action scores independently from the person box and the secondary box before summing them. As we can see from the poor results of our bounding-box-only model (second line of Table~\ref{table:hico_results}b), such prediction is hard.


With our fusion networks, we immediately see improvements over the full-image network (Table~\ref{table:hico_results}c). The weighted loss, which penalizes mistakes on positive examples more heavily as described in Section~\ref{subsec:MIL}, helps push the mAP higher by about 2.5 mAP for both our networks. The Fusion-2 network, which performs dimensionality reduction before local and global feature concatenation, has a slight edge probably due to lower number of parameters (Fusion-1 has $1024\times 512$ parameters for dimensionality reduction and Fusion-2 has $2\times 512\times 256$, lesser by a factor of 2).


\begin{table}[h!]
	\begin{center}
		\begin{tabular}{l|c}
			\hline
			\multicolumn{1}{c|}{\bf Method} 			& {\bf mAP }	\\ \hline
			Dense Trajectory + Pose~\cite{pishchulin2014fine}					& 5.5		\\ 
			VGG-16, R$^\ast$CNN~\cite{gkioxari2015rstarcnn}						& 26.7		\\ \hline
            Fusion-1, label per ground truth person instance				& 32.06		\\ 
			Fusion-2, label per ground truth person instance				& {\bf 32.24} \\
			Fusion-1, MIL over ground truth person instances				& 31.68		\\ 
			Fusion-2, MIL over ground truth person instances				& 31.89		  \\ 
 \hline
			Fusion-2, label per detected person instance				& 32.02		  \\
            Fusion-2, MIL over detected person instances				& 31.81		  \\

		\end{tabular}
	\end{center}
	\caption{Results on the MPII test set (obtained by submitting our output files by email to the authors of~\cite{pishchulin2014fine}).}
	\label{table:mpii_results}
\end{table}

{\bf MPII Results.} 
On the MPII dataset, we compare our networks with previously published baselines from Pischulin \emph{et al.}~\cite{pishchulin2014fine} and Gkioxari \emph{et al.}~\cite{gkioxari2015rstarcnn}. Our networks are trained with a learning rate of $10^{-4}$ with a decay of 0.1 every 12000 iterations, for 40000 iterations. We only finetune till the $fc6$ layer due to the smaller amount of training data than in HICO. We do not use the weighted loss on this dataset, as we did not find it to make a difference.

Table~\ref{table:mpii_results} shows the MPII results. The trend is similar to that in Table \ref{table:hico_results}: our fusion networks outperform previous methods, with Fusion-2 having a lead over Fusion-1. Recall that the MPII training set comes with ground truth person instances, which gives us a chance to examine the effect of MIL. If we assume that the assignment of labels to the people in the image is unknown and use the MIL framework, we see a small dip in performance as opposed to assuming that the label applies to each person in the image (last two rows of Table~\ref{table:mpii_results}). The latter gives us more training data along with full supervision and improves over MIL by around 0.4 mAP. We also tried training the network with detected person bounding boxes instead of groundtruth boxes and found that the performance was very similar, indicating that groundtruth boxes may not be necessary if there is no ambiguity in assignment of labels.

\begin{figure}[t!]
    \centering
    \includegraphics[width=\linewidth]{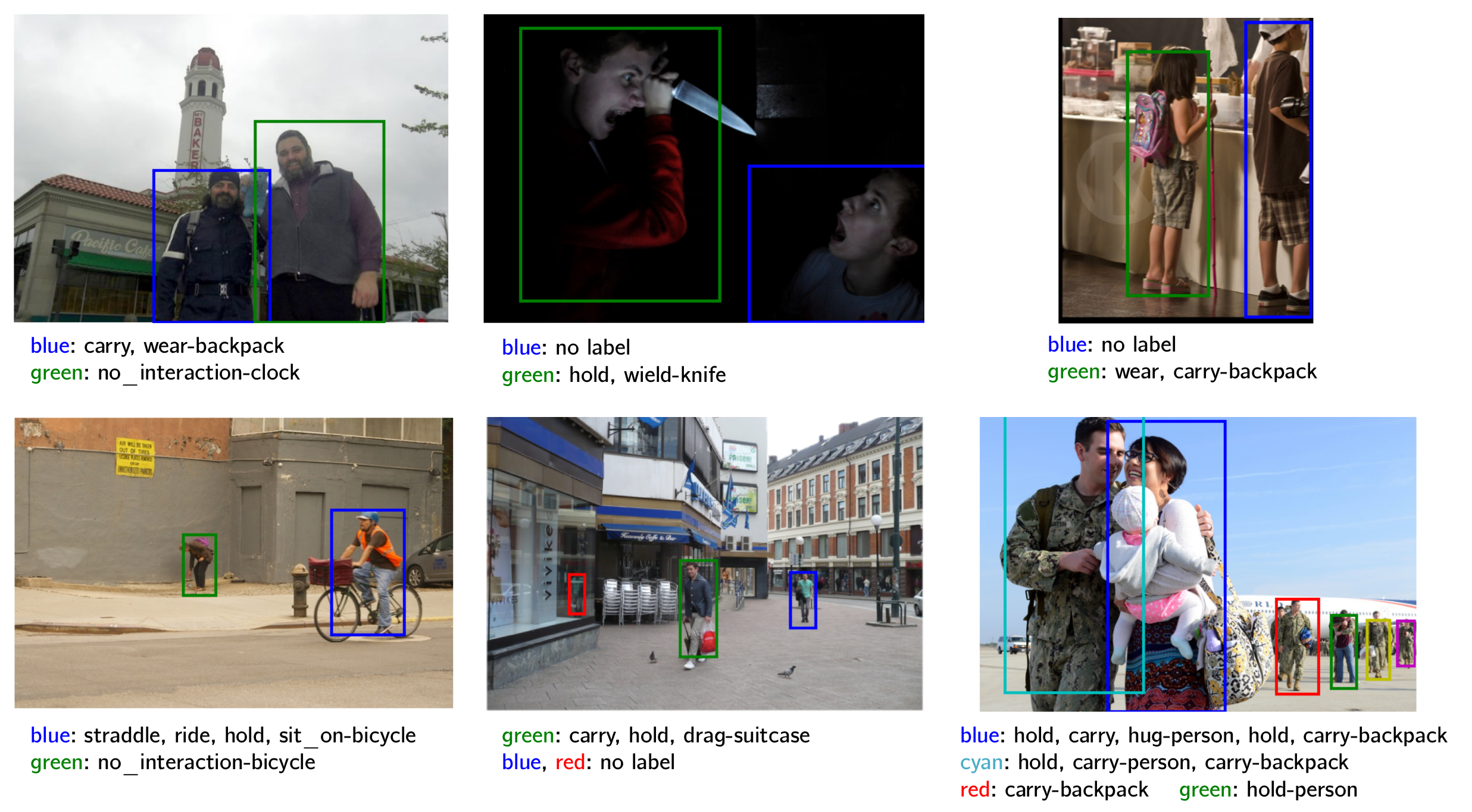}
    \caption{Predictions of our Fusion-2 model on the HICO test set. Detected person instances are marked in different colors and corresponding action labels are given underneath.}
    \label{fig:hico_mil_good}
\end{figure}

{\bf Qualitative Results.}
Figure~\ref{fig:hico_mil_good} displays some of the predictions of our best-performing network on the HICO dataset. In spite of the lack of explicit supervision of which labels map onto a specific person instance, the network learns to reasonably assign labels to the correct person instance. It is interesting to note a few minor mistakes made by the network: in the top left example, the network confuses the tower in the background for a clock tower, and assigns the label `no\_interaction-clock' to one of the people. In the middle example of the second row, there is a false person detection (marked in red) due to the reflection in the glass, but it does not get an activity prediction since the highest-scoring label has confidence less than 0.5. 

\begin{figure}[h!]
    \centering
    \includegraphics[width=\linewidth]{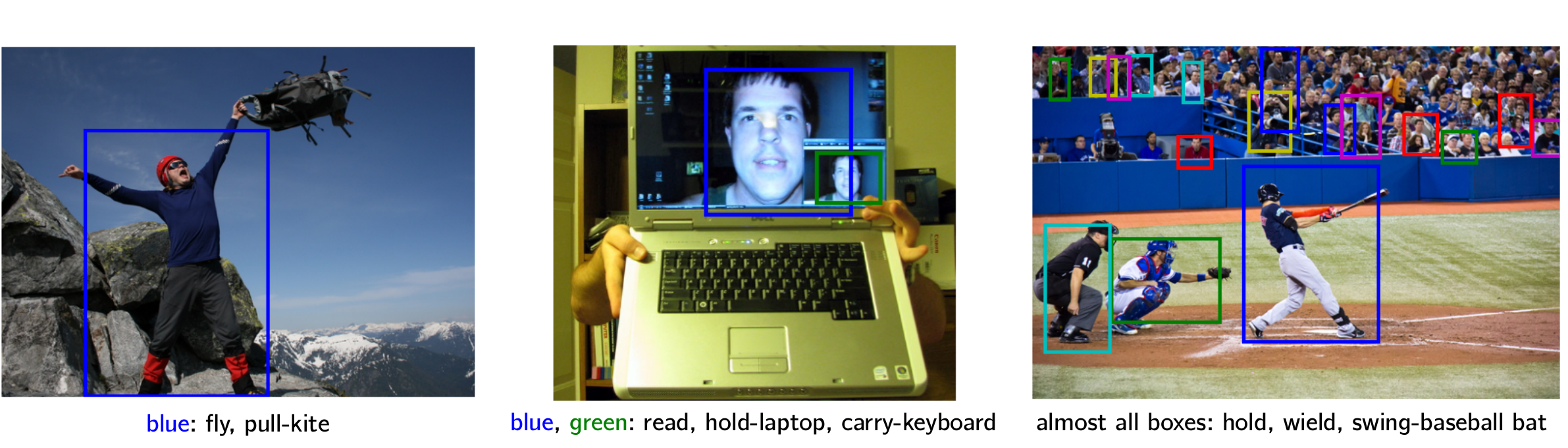}
    \caption{Failure examples on HICO. Incorrect classification of objects/actions, wrong interacting person detection, and inability to assign labels to correct person instances due to weak supervision and sampling are common issues.}
    \label{fig:hico_mil_bad}
\end{figure}

Figure~\ref{fig:hico_mil_bad} shows some of the failures of our system on the HICO dataset. Unusual use-cases of an object such as swinging around a backpack can confuse the deep network into misclassifying the object as in the leftmost image. Since our system relies on detected people, we can either miss or produce false positives, or label the wrong instances as shown in the middle image. Lastly, one drawback of the weakly supervised MIL framework is that it is unable to distinguish labels in a crowded scenario, especially when the crowd occurs only in specific settings such as sports games (right image).


\section{Visual Question Answering Results}
\label{subsec:mcq}

{\bf Dataset and Tasks.} In this section, we evaluate the performance of features extracted by our networks on two types of questions from the Madlibs dataset~\cite{yu2015visual} that specifically target people's activities and their interactions with objects. The first type, `Person's Activity,' asks us to choose an option that best describes the activity of the indicated person/people, while the second type, `Pair's Relationship,' asks us to describe the relationship between the indicated person and object(s). The indicated people and objects come from ground truth annotations on the MS COCO dataset~\cite{lin2014microsoft}, from which MadLibs is derived, so there is no need to perform any automatic detection. The prompt is fixed for all questions of a particular type: `The person/people is/are \underline{$\quad\quad$}' and `The person/people is/are \underline{$\quad\quad$} the object(s)'. 

The training data for MadLibs consists of questions paired with correct answers. 
There are 26528 and 30640 training examples for the activity and relationship questions, respectively (the total number of distinct images is only about 10K, but a single image can give rise to multiple questions centered on different person and object instances). In the test data, each question contains four possible answer choices, of which one is the correct answer (or best answer, in case of confusing options). Depending on the way the distractor options are selected, test questions are divided into two categories, Easy and Hard. The test sets for the activity and relationship types have 6501 and 7595 questions respectively, and each comes with Easy and Hard distractor options. Hard options are often quite confusing, with even humans disagreeing on the correct answer. Thus, the performance on filtered hard questions, on which human annotators agree with the `correct' answer at least $50\%$ of the times, is also measured. Since MadLibs does not provide a set of multiple choice questions for validation, we created our own validation set of Easy questions by taking 10\% of the training images and following the distractor generation procedure of~\cite{yu2015visual}.

{\bf Models and Baselines.}
Similarly to~\cite{yu2015visual}, we use normalized Canonical Correlation Analysis (nCCA)~\cite{gong2014multi} to learn a joint embedding space to which the image and the choice features are mapped. Given a question, we select the choice that has the highest cosine similarity with the image features in the joint embedding space as the predicted answer.

On the text side, we represent each of the choices by the average of the 300-dimensional word2vec features~\cite{mikolov2013distributed} of the words in the choice. In the case that a word is out of the vocabulary provided by~\cite{word2vec}, we represent it with all zeros. 

On the image side, we compare performance obtained with three types of features. The first is obtained by passing the entire image, resized to $224\times 224$ pixels, through the vanilla (ImageNet-trained) VGG-16 network and extracting the $fc7$ activations. This serves as the baseline, similar to the original work of Yu \emph{et al.}~\cite{yu2015visual}. The second type of feature is obtained by passing the entire image through our activity prediction network that uses full image inputs. We compare both the $fc7$ activations (of length 4096) and the class label activations (of length 600). The third type of feature is extracted by our Fusion-2 architecture (as detailed in Section~\ref{sec:method}). As our MadLibs question types target one or more specific people in the image, we feed in the person bounding boxes as ROIs to our network (for the relationship questions, we ignore the object bounding box). In the case that a particular question targets multiple people, we perform max pooling over the class label activations of the distinct people to obtain a single feature vector. Note that we found it necessary to use the class label activations before passing them through the logistic sigmoid/softmax as the squashing saturated the scores too close to 0 or 1. 

To train the nCCA model, we used the toolbox of Klein et al.~\cite{klein2015associating}. We set the CCA regularization parameter using the validation sets we created, resulting in values of 0.01 and 0.001 for the $fc7$ and class score features respectively. Our learned nCCA embedding space has dimensionality of 300 (same as the dimensionality of word2vec).

\begin{table}[h!]
	\begin{center}
		\begin{tabular}{ll|ccc|ccc}
            \hline
			\multicolumn{2}{c|}{} & \multicolumn{3}{|c|}{\bf Person's Activity} & \multicolumn{3}{c}{\bf Pair's Relationship}\\ 
			\multicolumn{1}{c}{\bf Dataset:Network} \ &- \ {\bf Feature} & {\ Easy\ } & {\ Hard\ } & {\ Fil. H.\ } & {\ Easy\ } & {\ Hard\ } & {\ Fil. H. \ }\\ \hline
			\multicolumn{1}{l}{ImageNet:VGG-19~\cite{yu2015visual}} \ &- \ fc7  & 80.7 & 65.4 & 68.8 & 63.0 & 54.3 & 57.6 \\ 
			\multicolumn{1}{l}{ImageNet:VGG-16} \ &- \ fc7  			& 80.79 & 65.14 & 67.73 & 71.45 & 51.47 & 56.28 \\ \hline
			\multicolumn{1}{l}{HICO:VGG-16, Full Im.}\ 	 &- \ cls\_score 		& 86.03 & 68.74 & 72.06 & 77.25 & 54.10 & 59.77 \\ 
			\multicolumn{1}{l}{HICO:VGG-16, Full Im.}\ 	 &- \ fc7 				& 86.54 & 69.14 & 72.39 & 77.96 & 55.76 & 61.03\\ \hline
			\multicolumn{1}{l}{HICO:Fusion-2}\   &- \ cls\_score 	& 86.66 & 70.05 & 73.46 & 78.29 & 55.52 & 61.39 \\ 
			\multicolumn{1}{l}{MPII:Fusion-2}\   &- \ cls\_score 	& 83.23	& 68.11	& 70.89 & 72.81 & 52.75 & 57.68 \\ \hline
			\multicolumn{1}{l}{HICO+MPII:Fusion-2}\   &- \ cls\_score 	& {\bf 87.57} & {\bf 71.13} & {\bf 74.45} & {\bf 78.50} & {\bf 56.17} & {\bf 62.06}

        \end{tabular}
	\end{center}
	\caption{Performance of different visual features on Activity and Relationship MadLibs questions (Fil. H. $\equiv$ Filtered Hard). See text for discussion.}
	\label{table:madlibs}
\end{table}

{\bf Question Answering Performance.}
The first two rows of Table~\ref{table:madlibs} contain the accuracies from the vanilla VGG baseline of Yu et al.~\cite{yu2015visual} and our reproduction. Some of our numbers deviate from those of~\cite{yu2015visual}, probably owing to the different features used (VGG-16 v/s VGG-19), CCA toolboxes, and hyperparameter selection procedures. 
From the second row of Table~\ref{table:madlibs}, using the vanilla VGG features gives an accuracy of $80.79\%$ and $71.45\%$ on the Easy Person Activity and Easy Pair Relationship questions respectively. By extracting features from our full-image network trained on the HICO dataset, we obtain gains of around $6-7\%$ on the Easy questions (rows 3-4). It is interesting to note that the 600-dimensional class label features give performance comparable to the 4096-dimensional $fc7$ features. Next, features from our Fusion-2 network trained on HICO (row 5) help improve the performance further. The Fusion-2 network trained on the smaller MPII dataset (row 6) gives considerably weaker performance. Nevertheless, we obtain our best performance by concatenating class label predictions from both HICO and MPII (last row of Table \ref{table:madlibs}), since some of the MPII categories are complementary to those of HICO, especially in the cases when a person is not interacting with any object. Compared to our baseline (row 2), we obtain an improvement of $6.8\%$ on the Easy Activity task, and $7.5\%$ on the Easy Relationship task. For the Hard Activity task, our improvements are $6\%$ and $6.7\%$ on the unfiltered and filtered questions, and for the Hard Relationship task, our improvements are $4.7\%$ and $5.8\%$ on the unfiltered and filtered questions respectively.

\begin{figure}[h!]
    \centering
    \includegraphics[width=\linewidth]{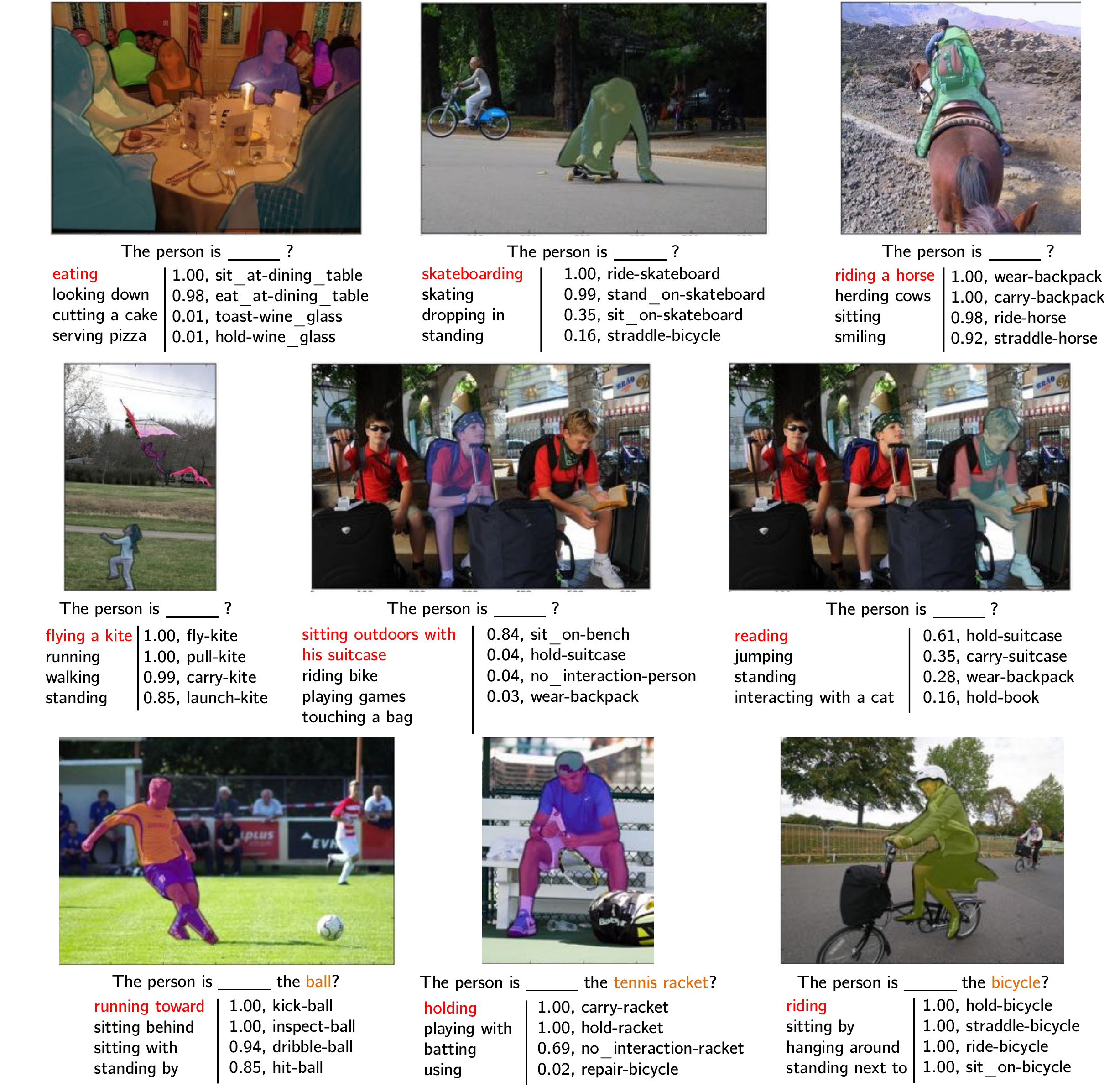}
    \caption{Correctly answered questions of the person activity type (first two rows) and person-object relationship type (last row). The subjects of the questions are highlighted in each image. The left column below each image shows the answer choices, with the correct choice marked in red. The right column shows the activity labels and scores predicted by our best network.}
    \label{fig:madlibs_good}
\end{figure}

{\bf Qualitative Results.} Figure~\ref{fig:madlibs_good} shows a range of correctly answered multiple choice questions using our best-performing features. 
By examining top labels predicted by our network, we can gain intuition into the choices of our model as these are easily interpretable unlike $fc7$ features of the VGG-16 network. In fact, our top predicted labels often align very closely to the correct answer choice. In the top left image of Figure~\ref{fig:madlibs_good}, the question targets multiple people and the label scores max pooled over the people correctly predict the activity of sitting at and eating at the dining table. In the middle image of the first row, the question targets the skateboarder. Accordingly, our network gives a high score for skateboard-related activities, and a much lower score for the bicyclist in the background. In the rightmost image in the first row, our network also correctly predicts the labels for `ride, straddle-horse' along with `wear, carry-backpack' (which is not one of the choices). The middle and right images in the middle row show that our predictions change depending on the target bounding box: the `hold-book' label has a much higher probability for the boy on the right, even though the network was trained using weak supervision and MIL, as detailed in Section~\ref{sec:method}. 

\begin{figure}[t!]
    \centering
    \includegraphics[width=\linewidth]{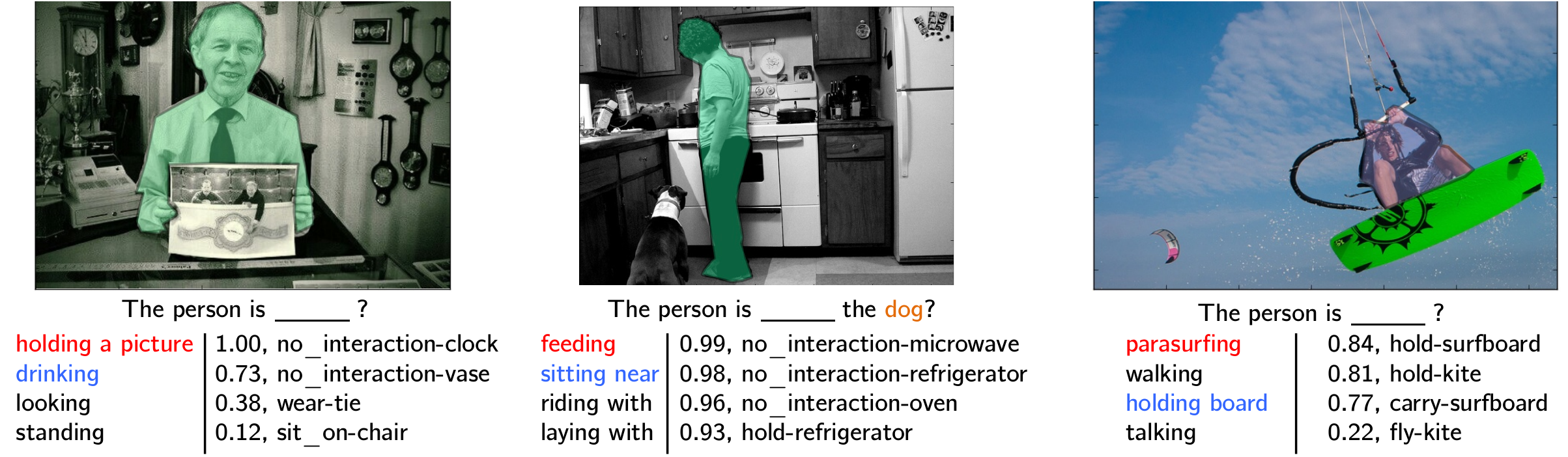}
    \caption{Failure examples. The correct choice is marked in red, and the predicted answer in blue. Failure modes mainly belong to three classes as illustrated (left to right): correct predictions but unfamiliar object (`picture'); incorrect predictions (`dog' missed); and a mix of the first two, i.e., partly correct predictions and unfamiliar setting.}
    \label{fig:madlibs_bad}
\end{figure}

Figure~\ref{fig:madlibs_bad} displays some of the common failure modes of our system. In the leftmost image, even though the predicted activity labels are correct, the target object of the question (`picture') is absent from the HICO and MPII datasets so the labels offer no useful information for answering the question. The network can also make wrong predictions, as in the middle image. In the rightmost image, the choices are rather hard and confusing as the person is indeed holding onto a kite as well as a surfboard in an activity best described as `parasurfing' or `windsurfing'.


\section{Conclusion}
\label{sec:conclusion}

In this paper, we developed effective models exploiting local and global context to make person-centric activity predictions and showed how Multiple Instance Learning could be used to train these models with weak supervision. Even though we used a simple global contextual representation, we obtained state-of-the-art performance on two different datasets, outperforming more complex models like R$^*$CNN. In future work, we hope to further explore more sophisticated contextual models and find better ways to train them on our target datasets, which feature hundreds of class labels with highly unbalanced label distributions.

We have also shown how transferring the knowledge from models trained on specialized activity datasets can improve performance on VQA tasks. While we demonstrated this on fairly narrow question types, we envision a more general-purpose system that would have access to many more input features such as person attributes, detected objects, scene information, etc.\ and appropriately combine them based on the question and image provided. 

\bibliographystyle{splncs}
\bibliography{egbib}
\end{document}